# Local Loss Optimization in Operator Models: A New Insight into Spectral Learning

**Borja Balle**  BBALLE@LSI.UPC.EDU
**Ariadna Quattoni**  AQUATTONI@LSI.UPC.EDU
**Xavier Carreras**  CARRERAS@LSI.UPC.EDU
Universitat Politècnica de Catalunya, Barcelona

## Abstract

This paper re-visits the spectral method for learning latent variable models defined in terms of observable operators. We give a new perspective on the method, showing that operators can be recovered by minimizing a loss defined on a finite subset of the domain. This leads to a derivation of a non-convex optimization similar to the spectral method. We also propose a regularized convex relaxation of this optimization. In practice our experiments show that a continuous regularization parameter (in contrast with the discrete number of states in the original method) allows a better trade-off between accuracy and model complexity. We also prove that in general, a randomized strategy for choosing the local loss succeeds with high probability.

## 1. Introduction

Structured latent variable models (e.g. Hidden Markov Models or Hidden Conditional Random fields) have become an essential modelling tool in multiple areas of machine learning such as Computer Vision, Natural Language Processing, and Bioinformatics. The power of these models resides in their ability to explain dependences in observed data using hidden unobserved variables. However, this expressivity comes at a cost: in general inducing the parameters of the model from observed data is computationally hard. In practice, despite the intrinsic difficulty, powerful heuristic methods have been developed. Most of these methods can be interpreted as instances of the Expectation–Maximization algorithm (Dempster et al., 1977). EM is an iterative algorithm that tries to minimize a non-convex objective function. One of its appeals is that it carries an intuitive interpretation, i.e. it minimizes the empirical error over a set of observed sequences. Its drawback is that since it attempts to minimize a non-convex function it is suceptible to local optima issues.

Recently a new line of work on learning structured latent variable models has emerged. It is the so-called *spectral learning method*, introduced by (Hsu et al., 2009) in the context of HMM and also applied to many other models such as Reduced Rank HMM (Siddiqi et al., 2010), Kernelized HMM (Song et al., 2010), Predictive State Representations (Boots et al., 2011), Latent Tree Graphical Models (Parikh et al., 2011), Finite States Transducers (Balle et al., 2011), and Quadratic Weighted Automata (Bailly, 2011). This method dodges the two main drawbacks of the EM algorithm: it always finds a global optimum, and its running time is linear in the number of training examples. The key insight of the spectral approach is to represent the distribution computed by the model in terms of observable operators and show that (under certain assumptions) two models with similar operators compute *similar* functions (under some metric).

The learning method then provides a set of equations, involving statistics computed from data, from which operators can be induced by computing approximate regularized solutions. In particular, all the works cited above share a common ingredient: the use of a Singular Value Decompostion for obtaining operators; hence the name spectral method. One of the appeals of this approach is that in general it can be rigorously studied using sensitivity analysis to bound the effect of perturbations on the equations used to recover operators from data. Altogether, it seems fair to assert that some of the theoretical aspects of the spectral method are now well understood. When contrasted with EM, there is little doubt that the spectral method is a very attractive alternative. However, EM seems to have some advantages on the eye of the researchers interested in exploiting latent variable models for a given application. Namely, its generic nature makes it easier to apply to new models and applications.

We believe some important aspects that can ease the applicability of spectral methods to real world problems have been overlooked in previous analysis. Some of these issues are addressed in the present paper.

Our first contribution is to re-visit the problem of learning observable operators from a loss minimization perspective. In particular, we give a formulation of the problem in terms of a regularized *local* loss minimization. We emphasize the local aspect of this minimization – which means that, in order to learn a function computed by an operator model, it is enough to observe its behavior on a finite set of elements. This is in contrast to the global loss formulation used in iterative algorithms such as EM.

To solve the local loss minimization we derive two optimization algorithms. The first algorithm frames the problem as minimizing a non-convex local loss function. We show that under certain conditions the standard SVD method can be seen as an optimizer for this objective.

Our second contribution is to propose a regularized convex relaxation of the local loss minimization. A feature of the SVD method is that only one discrete parameter, the *number of states*, needs to be tuned. In practice, this means that the space of all possible hypothesis can be exhaustively explored in relatively short time. However, in many cases, tuning this coarse-grained parameter is not enough for attaining an optimal trade-off between empirical error and model complexity. In contrast, our convex optimization algorithm takes a continuous regularization parameter that can be tuned in order to achieve an optimal trade-off. In practice, our synthetic experiments show that our method can be more robust to some spectral properties of the target distribution that represent a challenge for the SVD method.

In this paper we also address another important practical issue overlooked by previous work. In the general case, the consistency of both the SVD method and our optimization algorithms depend on a rather strong hypothesis. Namely, that the "correct" subset of the domain where the operators must be optimized – in our terms, the local loss function – is known to the algorithm. Though very convenient in theoretical studies, in practice this assumption does not seem very realistic.

Our third contribution is to prove that a simple randomized strategy can identify a correct local loss function with high probability. More precisely, we give bounds on the number of examples required by our loss-selection algorithm that depend polynomially on some parameters of the target.

We choose to present our results in the setting of Weighted Automata. This framework encompasses several of the models considered in the literature on the spectral method: HMM, reduced-rank HMM, PNFA, QWA and rational stochastic languages. With some modifications, this framework can also deal with input-output models like FST and PSR. In general, models defined in terms of a finite state machines over some finite alphabet can be formulated using Weighted Automata.

## 2. Weighted Automata and Hankel Matrices

### 2.1. Preliminaries and Notation

Let $\Sigma$ be a finite alphabet with $m$ symbols. We write $\Sigma^*$ for the set of all strings over $\Sigma$ and use $\lambda$ to denote the empty string. The *Hankel matrix* of a function $f : \Sigma^* \to \mathbb{R}$ over strings is a bi-infinite matrix $H_f : \Sigma^* \times \Sigma^* \to \mathbb{R}$ with its entries indexed by prefixes and suffixes: $H_f(u,v) = f(uv)$. The *rank* of $f$ is defined as $\mathrm{rank}(f) = \mathrm{rank}(H_f)$, which may in principle be infinite. Given sets of prefixes and suffixes $\mathcal{U}, \mathcal{V} \subset \Sigma^*$, we define the *Hankel sub-block* $H : \mathcal{U} \times \mathcal{V} \to \mathbb{R}$ of $H_f$ as $H(u,v) = f(uv)$. Note that when $|\mathcal{U}| = p$, $|\mathcal{V}| = s$ we have $H \in \mathbb{R}^{p \times s}$. In general, given $\mathcal{U}$ and $\mathcal{V}$ one has $\mathrm{rank}(H) \leq \mathrm{rank}(H_f)$. We say that the pair $(\mathcal{U}, \mathcal{V})$ is a *basis* for $f$ if $\mathrm{rank}(H) = \mathrm{rank}(H_f)$. Note that then it must be the case that $p, s \geq rank(H_f)$. For any symbol $a \in \Sigma$, we also define the sub-block $H_a \in \mathbb{R}^{p \times s}$ as $H_a(u,v) = f(uav)$.

A *weighted automata* (WA) over $\Sigma$ with $n$ states is a tuple $A = \langle \alpha_1^\top, \alpha_\infty, \{A_a\}_{a \in \Sigma} \rangle$, where $\alpha_1, \alpha_\infty \in \mathbb{R}^n$, $A_a \in \mathbb{R}^{n \times n}$. We write $|A|$ for the number of states of $A$. The function $f_A : \Sigma^* \to \mathbb{R}$ defined by $A$ is given by

$$f_A(x_1 \cdots x_t) = \alpha_1^\top A_{x_1} \cdots A_{x_t} \alpha_\infty = \alpha_1^\top A_x \alpha_\infty \ . \quad (1)$$

It is obvious from the definition that if $M \in \mathbb{R}^{n \times n}$ is an *invertible* matrix, the WA $B = \langle \alpha_1^\top M, M^{-1} \alpha_\infty, \{M^{-1} A_a M\} \rangle$ satisfies $f_B = f_A$. Sometimes $B$ is denoted by $M^{-1} A M$.

A probability distribution $\mathcal{D}$ over $\Sigma^*$ receives the name of a *stochastic languange*. We say that $\mathcal{D}$ has *full support* if $\mathcal{D}(x) > 0$ for all $x \in \Sigma^*$. A stochastic language is *rational* if there exists a WA $A$ such that

$f_A(x) = \mathcal{D}(x)$ for all $x \in \Sigma^*$.

By default all vectors are assumed to be columns. The Moore–Penrose pseudo-inverse of a matrix $M$ is denoted by $M^+$. A *rank factorization* of a matrix $M \in \mathbb{R}^{m \times n}$ with $\text{rank}(M) = r$ is a pair $Q \in \mathbb{R}^{m \times r}$, $R \in \mathbb{R}^{r \times n}$ such that $M = QR$ and $\text{rank}(Q) = \text{rank}(R) = r$. We denote the $i$th row of $M$ by $M(i,:)$, and the $j$th column by $M(:,j)$. For Hankel matrices and Hankel sub-blocks, rows and columns are respectively indexed by prefixes and suffixes. The notation $\|\cdot\|$ is used for the $\ell_2$ norm of vectors and matrices. Similarly, $\|\cdot\|_F$ denotes the *Frobenius* norm, and $\|\cdot\|_*$ the *nuclear* norm.

## 2.2. Probability Distributions over Strings

Throughout the paper it is assumed that some sub-blocks of the Hankel matrix $H_f$ are known, either exactly or in an approximate form. Obviously, in practice it only makes sense to consider targets for which (approximations of) these sub-blocks can be effectively obtained, say by examples drawn from a probility distribution, say by making queries to some oracle. In general, most spectral methods discussed in Section 1 are used for learning probability distributions defined by some form of finite state machine. In these cases, the entries of $H$ are probabilities and usually a sample drawn from the corresponding distribution is used for obtaining empirical estimates of this probabilities, yielding an approximate Hankel sub-block $\widehat{H}$.

Though we shall not fix any particular probabilistic model, it is worth noting that our results apply seamlessly to most of the settings cosidered so far. In particular, we can deal with the following two settings: when $f$ defines probabilities over finite prefixes (like in the HMM formulations) and words are sampled from these distributions conditioned on an externally (fixed or randomly) given length; and, when $f$ is a rational stochastic language. Furthermore, in the latter case our model encompasses the settings where a sample is used to estimate probabilities of words $f(x)$, prefixes $f(x\Sigma^*)$, or substrings $f(\Sigma^* x \Sigma^*)$, since it is not difficult to see that when $f$ is given by some WA with $n$ states, there exists another WA with $n$ states computing prefix and substring probabilities (Luque et al., 2012).

## 2.3. Duality between WA and Factorizations

Let $f : \Sigma^* \to \mathbb{R}$ be a function over strings with Hankel matrix $H_f$. We recall the following result (see (Beimel et al., 2000)): $\text{rank}(f) = r < \infty$ if and only if $f = f_A$ for some WA $A$ with $r$ states and for any WA $A$ such that $f_A = f$ then $|A| \geq r$. If $f = f_A$ and $|A| = \text{rank}(f)$ we say that $A$ is *minimal* for $f$.

Our ultimate goal is to learn a function $f : \Sigma^* \to \mathbb{R}$ of finite rank by observing a sub-block of its Hankel matrix. Since our hypotheses will be functions computed by weighted automata, a natural question to ask is the relation between (minimal) WA for $f$ and sub-blocks of $H_f$. Our first observation is that any minimal WA for $f$ induces a "nice" factorization of any sub-block $H$ defined on a "good" set of prefixes and suffixes.

Let $A$ be a minimal WA for some $f$ of rank $r$. Then $A$ induces a rank factorization of the Hankel matrix of $f$ of the form $H_f = P_f S_f$, where $P_f \in \mathbb{R}^{\infty \times r}$ and $S_f \in \mathbb{R}^{r \times \infty}$ are defined as: $P_f(u,:) = \alpha_1^\top A_u$, and $S_f(:,v) = A_v \alpha_\infty$. Actually, for any sets of prefixes $\mathcal{U}$ and suffixes $\mathcal{V}$, $A$ also induces a factorization $H = PS$ of the associated sub-block with $P \in \mathbb{R}^{p \times r}$ and $S \in \mathbb{R}^{r \times s}$. Furthermore, we can show that if $(\mathcal{U}, \mathcal{V})$ is a basis of $f$, then $H = PS$ is a rank factorization. Indeed, the inequalities $r = \text{rank}(H) \leq \min\{\text{rank}(P), \text{rank}(S)\}$ and $p, s \geq r$, imply that $\text{rank}(P) = \text{rank}(S) = r$. Note that from $P(u,:) = \alpha_1 A_u$ and $S(:,v) = A_v \alpha_\infty$ one can also derive the following useful factorization: $H_a = PA_a S$.

Thus, we have seen how a minimal WA for $f$ induces a rank factorization of $H$ provided that $\mathcal{U}$ and $\mathcal{V}$ form a basis of $f$. The following lemma shows that this relation can be reversed. Together, these two facts show that minimal WA for $f$ and rank factorizations of $H$ are "dual" whenever $(\mathcal{U}, \mathcal{V})$ is a basis.

**Lemma 1.** *Suppose $(\mathcal{U}, \mathcal{V})$ is a basis of $f$ with $\lambda \in \mathcal{U}$ and $\lambda \in \mathcal{U}$. Let $h_{r,\lambda}^\top = H(\lambda,:)$ and $h_{c,\lambda} = H(:,\lambda)$ be the respective row and column of $H$ associated with $\lambda$. For any rank factorization $H = QR$, let $A = \langle \alpha_1^\top, \alpha_\infty, \{A_a\} \rangle$ be the WA given by: $\alpha_1^\top = h_{r,\lambda}^\top R^+$, $\alpha_\infty = Q^+ h_{c,\lambda}$, and $A_a = Q^+ H_a R^+$. Then $A$ is a minimal WA for $f$.*

*Proof.* Let $B = \langle \beta_1, \beta_\infty, \{B_a\} \rangle$ be a minimal WA for $f$ inducing a rank factorization $H = PS$. It suffices to prove that there exists an invertible $M$ such that $A = M^{-1}BM$. Let $M = SR^+$. Since $(Q^+P)(SR^+) = Q^+ HR^+ = I$, we see that $M$ is invertible with inverse $M^{-1} = Q^+ P$. Now we check that the operators of $A$ correspond to the operators of $B$ under the change of basis $M$. First, we see that $A_a = Q^+ H_a R^+ = Q^+ P B_a S R^+ = M^{-1} B_a M$. Now observe that by the definitions of $S$ and $P$ we have $\beta_1^\top S = h_{r,\lambda}^\top$ and $P\beta_\infty = h_{c,\lambda}$. Thus, we see that $\alpha_1^\top = \beta_1^\top M$ and $\alpha_\infty = M^{-1} \beta_\infty$. $\square$

From now on, we assume without loss of generality that any basis $(\mathcal{U}, \mathcal{V})$ contains the empty string $\lambda$ as a prefix and a suffix.

The spectral algorithm of (Hsu et al., 2009) can be easily derived using Lemma 1. Basically, it accounts to taking the rank factorization $H = (HV)V^\top$, where $H = U\Lambda V^\top$ is a compact SVD. In next section we will derive another algorithm based on loss minimization that yields similar results.

## 3. Learning WA via Loss Minimization

In spirit, our algorithm is similar to the spectral method in the sense that in order to learn a function $f : \Sigma^* \to \mathbb{R}$ of finite rank, the algorithm infers a WA using (approximate) information from a sub-block of $H_f$. The sub-block used by the algorithm is defined in terms of a set of prefixes $\mathcal{U}$ and suffixes $\mathcal{V}$. Throughout this section we assume that $f$ is fixed and has rank $r$, and that a basis $(\mathcal{U}, \mathcal{V})$ of $f$ is given. How to find these sets of prefixes and suffixes given a sample is discussed in Section 4.

We state our algorithm under the hypothesis that sub-blocks $H$ and $\{H_a\}_{a \in \Sigma}$ of $H_f$ are known exactly. It is trivial to modify the algorithms to work in the case when only *approximations* $\widehat{H}$ and $\{\widehat{H}_a\}_{a \in \Sigma}$ of the Hankel sub-blocks are known.

For $1 \leq n \leq s$ we define the *local loss* function $\ell_n(X, \beta_\infty, \{B_a\})$ on variables $X \in \mathbb{R}^{s \times n}$, $\beta_\infty \in \mathbb{R}^n$ and $B_a \in \mathbb{R}^{n \times n}$ for $a \in \Sigma$ as:

$$\ell_n = \|HX\beta_\infty - h_{c,\lambda}\|_2^2 + \sum_a \|HXB_a - H_aX\|_F^2 \quad (2)$$

The operator learning algorithm is a constrained minimization of the local loss:

$$\min_{X, \beta_\infty, \{B_a\}} \ell_n(X, \beta_\infty, \{B_a\}) \text{ s.t. } X^\top X = I \quad \text{(SO)}$$

Intuitively, this optimization tries to *jointly* solve the optimizations solved by SVD and pseudo-inverse in the spectral method based on Lemma 1. In particular, likewise for the SVD-based method, it can be shown that (SO) is consistent whenever a large enough guess for $n$ is provided.

**Theorem 2.** *Suppose $n \geq r$. Then, for any optimal solution $(X^*, \beta_\infty^*, \{B_a^*\})$ to problem (SO), the weighted automata $B^* = \left\langle h_{r,\lambda}^\top X^*, \beta_\infty^*, \{B_a^*\} \right\rangle$ satisfies $f = f_{B^*}$*

The proof of this theorem is sketched in Appendix A.1. Though the proof is relatively simple in the case $n = r$, it turns out that the case $n > r$ is much more delicate – unlike in the SVD-based method, where the same proof applies to all $n \geq r$.

Of course, if $H$ and $\{H_a\}$ are not fully known, but approximations $\widehat{H}$ and $\{\widehat{H}_a\}$ are given to the algorithm, we can still minimize the *empirical local loss* $\widehat{\ell}_n$ and build a WA from the solution using the same method of Theorem 2.

Despite its consistency, in general the optimization (SO) is not algorithmically tractable because its objective function is quadratic non-positive semidefinite and the constraint on $X$ is not convex. Nonetheless, the proof of Theorem 2 shows that when $H$ and $\{H_a\}$ are known exactly, the SVD method can be used to efficiently compute an optimal solution of (SO). Furthermore, the SVD method can be regarded as an approximate solver for (SO) with an empirical loss function $\widehat{\ell}_n$ as follows. Find first an $\widehat{X}$ satisfying the constraints using the SVD of $\widehat{H}$, and then compute $\widehat{\beta}_\infty$ and $\{\widehat{B}_a\}$ by minimizing the loss (2) with fixed $\widehat{X}$ – note that in this case, the optimization turns out to be convex.

From this perspective, the bounds for the distance between operators recovered with full and approximate data given in several papers about the spectral method, can be restated as a sensitivity analysis of the optimization solved by the spectral algorithm. In fact, a similar analysis can be done for (SO), though we shall not pursue this direction here.

Instead, we shall present a convex relaxation of (SO) that addresses a practical issue in this optimization algorithm. That is, the fact that the only parameter a user can adjust in (SO) in order to trade accuracy and model complexity is the number of states $n$. Though the discreteness of this parameter allows for a fast model selection scheme through a full exploration of the parameter space, in some applications one may be willing to invest some time in exploring a larger, more fine-grained space of parameters, with the hope of reaching a better trade-off between accuracy and model complexity. The algorithm presented in next section does this by incorporating a continuous regularization parameter.

### 3.1. A Convex Local Loss

The main idea in order to obtain a convex optimization problem similar to (SO) will be to remove $X$, since we have already seen that it is the only source of non-convexity in the optimization. However, the new convex objective will need to incorporate a term that enforces the optimization to behave in a similar way as (SO).

First note that the choice of $n$ effectively restricts the maximum rank of the operators $B_a$. Once this maximal rank is set, $X$ can be interpreted as enforcing a common "semantic space" between the different operators $B_a$ by making sure each of them works on a state space defined by the same projection of $H$. Furthermore, the constraint on $X$ tightly controls its norm and thus ensures that the operators $B_a$ will also have its norm tightly controlled to be in the order of $\|H_a\|/\|H\|$ – at least when $n = r$, see the proof of Theorem 2.

Thus, in order to obtain a convex optimization similar to (SO) we do the following. First, take $n = s$ and fix $X = I$, thus unrestricting the model class and removing the source of non-convexity. Then penalize the resulting objective with a convex relaxation of the term $\mathrm{rank}([B_{a_1}, \ldots, B_{a_m}])$, which makes sure the operators have low rank individually, and enforces them to work on a common low-dimensional state space.

More formally, for any regularization parameter $\tau > 0$, the *relaxed local loss* $\tilde{\ell}_\tau(B_\Sigma)$ on a matrix variable $B_\Sigma \in \mathbb{R}^{s \times ms}$ is defined as:

$$\tilde{\ell}_\tau = \|B_\Sigma\|_* + \tau \|HB_\Sigma - H_\Sigma\|_F^2 \ , \quad (3)$$

where we interpret $B_\Sigma = [B_{a_1}, \ldots, B_{a_m}]$ as a concatenation of the operators, and $H_\Sigma = [H_{a_1}, \ldots, H_{a_m}]$. Since $\tilde{\ell}$ is clearly convex on $B_\Sigma$, we can learn a set of operators by solving the convex optimization problem

$$\min_{B_\Sigma} \tilde{\ell}(B_\Sigma) \ . \quad (\mathrm{CO})$$

Given an optimal solution $B_\Sigma^*$ of (CO), we define a WA $B^* = \left\langle h_{r,\lambda}^\top, e_\lambda, B_\Sigma^* \right\rangle$, where $e_\lambda \in \mathbb{R}^s$ is the coordinate vector with $e_\lambda(\lambda) = 1$.

Some useful facts about this optimization are collected in the following proposition.

**Proposition 3.** *The following hold: (1) if $H$ has full column rank, then (CO) has a unique solution; (2) for $n = s$ and $\tau \geq 1$, the optimum value $\ell_s^*$ of (SO) and the optimum value $\tilde{\ell}_\tau^*$ of (CO) satisfy $\ell_s^* \leq \tilde{\ell}_\tau^*$; (3) suppose $\mathrm{rank}(H) = \mathrm{rank}([H_\Sigma, H])$ and let $[H_\Sigma, H] = U\Lambda[V_\Sigma^\top V^\top]$ be a compact SVD. Then, $B_\Sigma = (V^\top)^+ V_\Sigma^\top$ is a closed form solution for (CO) when $\tau \to \infty$*

*Proof.* Fact (1) follows from the observation that when $H$ has full rank the loss $\tilde{\ell}_\tau$ is strictly convex. For fact (2), suppose $B_\Sigma^*$ achieves the optimal value in (CO) and check that $\ell_s^* \leq \ell_s(I, e_\lambda, B_\Sigma^*) = \|HB_\Sigma^* - H_\Sigma\|_F^2 \leq \tilde{\ell}_\tau^*$. Fact (3) follows from Theorem 2.1 in (Liu et al., 2010) and the observation that when $\tau \to \infty$ optimization (CO) is equivalent to $\min_{B_\Sigma} \|B_\Sigma\|_*$ s.t. $HB_\Sigma = H_\Sigma$. $\square$

Note that in general approximations $\widehat{H}$ of $H$ computed from samples will have full rank with high probability. Thus, fact (1) tells us that either in this case, or when $p = n$, optimization (CO) has a unique optimum.

Furthermore, by fact (2) we see that minimizing the convex loss is also, in a relaxed sense, minimizing the non-convex loss which is known to be consistent. In addition, fact (3) implies that when $H$ has full rank and $\tau$ is very large, we recover the spectral method with $n = s$. These and other properties of (CO) appear in the experimens described in Section 5.

Optimization (CO) can be restated in several ways. In particular, by standard techniques, it can be shown that it is equivalent to a Conic Program on the intersection of a semi-definite cone (given by the nuclear norm), and a quadratic cone (given by the Frobenius norm). Similarly, the problem can also be fully expressed as a semi-definite program, though in general this conversion is believed to be inefficient. Altogether, the number of variables in (CO) is $ms^2$. Formulating the conic program yields $O(m^2s^2)$ varibles, and constraints in a space of size $O(mps + ms^2)$. When the fully semi-definite program is considered, the constraint space grows to dimension $O(m^2p^2s^2)$. This shows that finding a small basis, in particular, a basis defined over a small set of prefixes, is important in practice. We note here that the complexity of the SVD method scales similarly.

## 4. Choosing the Local Loss

We have already discussed why, in practice, it is important to have methods for finding a basis. In this section we show a fundamental result about basis. Namely, that simple randomized strategies for choosing a basis succeed with high probability. Furthermore, our result gives bounds on the number of examples required for finding a basis that depend polynomially on some parameters of the target function $f : \Sigma^* \to \mathbb{R}$ and the sampling distribution $\mathcal{D}$.

We begin with a well-known folklore result about the existence of minimal basis. This implies that in principle all methods for learning WA from sub-blocks of the Hankel matrix can work with a block whose size is only quadratic in the number of states of the target.

**Proposition 4.** *For any $f : \Sigma^* \to \mathbb{R}$ of rank $r$ there exists a basis $(\mathcal{U}, \mathcal{V})$ of $f$ with $|\mathcal{U}| = |\mathcal{V}| = r$.*

A WA $A = \langle \alpha_1^\top, \alpha_\infty, \{A_a\} \rangle$ is called *strongly bounded* if $\|A_a\| \leq 1$ for all $a \in \Sigma$. Note that this implies the boundeness of $f_A$ since $|f_A(x)| = |\alpha_1^\top A_x \alpha_\infty| \leq \|\alpha_1\| \|\alpha_\infty\|$. A function over strings $f$ of finite rank is called *strongly bounded* if there exists a strongly bounded minimal WA for $f$. Note that, in particular, all models of probabilistic automata discussed in Section 2.2 are strongly bounded.

Our result states that, under some simple hypothesis,

**Algorithm 1** Random Basis
   **Input:** strings $\mathcal{S} = (x^1, \ldots, x^N)$
   **Output:** basis candidate $(\mathcal{U}, \mathcal{V})$
   Initialize $\mathcal{U} \leftarrow \varnothing$, $\mathcal{V} \leftarrow \varnothing$
   **for** $i = 1$ **to** $N$ **do**
      Choose $0 \leq t \leq |x^i|$ uniformly at random
      Split $x^i = u^i v^i$ with $|u^i| = t$ and $|v^i| = |x^i| - t$
      Add $u_i$ to $\mathcal{U}$ and $v^i$ to $\mathcal{V}$
   **end for**

| $\sigma_n$ | | 50K | 100K | 150K | 200K |
|---|---|---|---|---|---|
| $\sim 10^{-1}$ | SVD | 0.0213 | 0.0137 | 0.0128 | **0.001** |
| | CO | **0.0199** | **0.0130** | **0.0121** | 0.0093 |
| $\sim 10^{-2}$ | SVD | 0.0501 | 0.0450 | 0.0426 | 0.0399 |
| | CO | **0.0460** | **0.0390** | **0.0362** | **0.0317** |
| $\sim 10^{-3}$ | SVD | 0.0310 | 0.0194 | 0.0186 | 0.0154 |
| | CO | **0.0259** | **0.0173** | **0.0181** | **0.0144** |

Table 1. Learning curves for three target distributions differing in the magnitude of the smallest singular value.

with high probability Algorithm 1 will return a correct basis when enough examples are examined.

**Theorem 5.** *Let $f : \Sigma^* \to \mathbb{R}$ be a strongly bounded function of rank $r$ and $\mathcal{D}$ a distribution over $\Sigma^*$ with full support. Suppose that $N$ strings sampled i.i.d. from $\mathcal{D}$ are given to Algorithm 1. Then, if $N \geq C\eta \log(1/\delta)$ for some universal constant $C$ and a parameter $\eta$ that depends on $f$ and $\mathcal{D}$, the output $(\mathcal{U}, \mathcal{V})$ is a basis for $f$ with probability at least $1 - \delta$.*

A proof of this result based on random matrix theory is given in Appendix A.

## 5. Experimental Results

We conducted synthetic and real experiments comparing the SVD and the Convex Optimization methods.

For the synthetic experiments, we created random PNFAs with alphabet sizes ranging from 2 to 10 symbols and a random number of states in the same range. For each random target model, we then sampled $k$ training sequences and trained models using SVD and CO. Results are reported in terms of $L_1$ error with respect to the true distribution (all results are averages of 10 sampling rounds). We fixed the set of prefixes and suffixes to be all substrings of length 1, following (Hsu et al., 2009). Table 1 shows learning curves for three target models. Each model was chosen randomly from a set of models that have the smallest singular value of $H$ in the same order of magnitude. For each target and method we show the error of the best model (i.e. optimal $n$ for the SVD method and optimal $\tau$ for the CO method). For the second distribution in Table 1, Figure 1.a shows the $L_1$ error of the CO method as a function of $\tau$. It also shows the error of the SVD method for different number of states.

Figure 1.b summarizes all results for the largest size of training set. For each target model, we show the average error of the two learned models as a function of the smallest singular value of the target. We observe that in general target models with smaller singular values are harder to learn, and it is in those cases that the CO approach obtains the largest gain in accuracy.

We also conducted experiments on natural language data, for the task of language modeling of syntactic part-of-speech tags (i.e. noun, verb, adjective, ...). This type of language models are a central building block in Natural Language Processing methods for tagging the words of a sentence with their syntactic function. We used the English Penn Treebank with a tag set of 12 symbols, and used the standard splits for training (39,832 sentences with avg. length of 23) and validation (1,700 sentences).

Figure 1.c plots curves on the validation set comparing the SVD method, the CO method, the standard EM algorithm, and two simple baselines based on statistics of single symbols (Unigram) and pairs of symbols (Bigram). For each model we plot the *word error rate* with respect to the nuclear norm of their operators. All hidden state models improve the baselines, while the CO method is able to improve over the SVD method. The EM method obtains the best error rates, though it is much slower to train (a factor of 100 times).

Finally, Table 2 shows the peformance of the SVD method using random sets of prefixes and suffixes. In this case, we generated substrings of up to 4 symbols, and sampled them according to their frequency on the training data. We used the random substrings to define Hankel sub-blocks of increasing dimensionalities. For each dimensionality, we trained a model using SVD, and chose the number of states that minimized error on validation data. Clearly, expanding the Hankel sub-block results in a benefit in terms of the error. For comparison, the table also reports the performance of EM with respect to the number of states.

## 6. Conclusion

In this paper we have attempted to facilitate the understanding and applicability of spectral approaches for learning weighted automata. In particular, we have made the following contributions: (1) formulate weighted automata learning as a local loss minimization; (2) show that under certain conditions the standard SVD approach is an optimizer of this local loss;

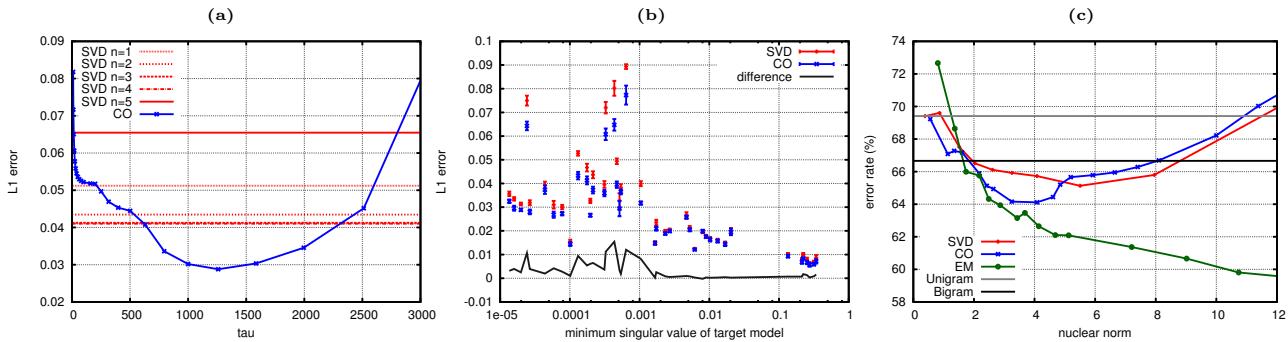

*Figure 1.* (a) Synthetic: the effect of the regularization parameter (tau) on a random distribution in terms of average $L_1$ error; for comparison, the plot also depicts the error of the SVD learner for different number of states. (b) Synthetic: comparison in terms of average $L_1$ error of SVD and CO learners on 40 random targets; the black line is the difference in errors for each target. (c) Part-of-speech models: error rates on the validation set with respect to the nuclear norm of models learned using SVD, CO and EM.

| | SVD | | | EM | |
|---|---|---|---|---|---|
| dim | n | error (%) | | n | error (%) |
| 12 | 8 | 65.1 | | 12 | 62.1 |
| 25 | 10 | 63.9 | | 15 | 61.4 |
| 50 | 12 | 63.2 | | 20 | 60.6 |
| 100 | 30 | 62.2 | | 30 | 59.4 |
| 300 | 38 | 61.6 | | 50 | 58.6 |
| 500 | 38 | 61.3 | | 100 | 57.9 |

*Table 2.* Error rates of the SVD method using increasing number of random prefixes and suffixes ($dim$); $n$ is the optimal number of states for each dimensionality. Error rates of EM are with respect to the number of states.

(3) propose a convex relaxation that permits fine tuning of the complexity–accuracy trade-off; (4) offer a provable correct method for estimating the scope of the local loss function from samples; and (5) show on synthetic experiments that under certain conditions the convex relaxation method is more robust than the SVD approach.

## A. Technical Proofs

### A.1. Proof Sketch for Theorem 2

The following two lemmas will be used in the proof.

**Lemma 6.** *Let $A = \langle \alpha_1^\top, \alpha_\infty, \{A_a\} \rangle$ be a WA with $n$ states. Suppose that $(\mathcal{U}, \mathcal{V})$ is a basis for $f_A$ and write $H = PS$ for the factorization induced by $A$ on this Hankel sub-block. For any $m$ and any pair of matrices $N \in \mathbb{R}^{m \times n}$ and $M \in \mathbb{R}^{n \times m}$ such that $PMN = P$, the WA $B = NAM = \langle \alpha_1^\top M, N\alpha_\infty, \{NA_aM\} \rangle$ satisfies $f_B = f_A$.*

**Lemma 7.** *Let $f : \Sigma^* \to \mathbb{R}$ be a function of finite rank $r$ and suppose that $(\mathcal{U}, \mathcal{V})$ is a basis for $f$. Then the matrix $H_\Sigma = [H_{a_1}, \ldots, H_{a_m}]$ has rank $r$.*

Now the following three facts can be established. To-gether, they imply the result.

*Claim 1:* The optimal value of problem (SO) is zero. Let $H = U\Lambda V^\top$ be a full SVD of $H$ and write $V_n \in \mathbb{R}^{s \times n}$ for the $n$ left singular vectors corresponding first $n$ singular values. Then consider the WA $A_n = \langle h_{r,\lambda}^\top V_n, (HV_n)^+ h_{c,\lambda}, \{(HV_n)^+ H_a V_n\} \rangle$ and show that $\ell_n(V_n, (HV_n)^+ h_{c,\lambda}, \{(HV_n)^+ H_a V_n\}) = 0$.

*Claim 2:* For any $n \geq r$, $A_n$ satisfies $f_{A_n} = f$. Apply Lemma 6 to show that $f_{A_n} = f_{A_r}$ for $n > r$, and then Lemma 1 to show $f_{A_r} = f$.

*Claim 3:* For any optimal solution $B^*$ one has $f_{B^*} = f_{A_n}$. Lemma 7 is used to show that $HX^*(X^*)^\top = H$. Then, Lemma 6 with $N = (X^*)^\top V_n$ and $M = V_n^\top X^*$ implies the claim.

### A.2. Proof of Theorem 5

We use the following result from (Vershynin, 2012).

**Theorem 8** (Corollary 5.52 in (Vershynin, 2012))**.** *Consider a probability distribution in $\mathbb{R}^d$ with full-rank covariance matrix $C$ and supported in a centered Euclidean ball of radius $R$. Also, let $\sigma_1 \geq \ldots \geq \sigma_d > 0$ be the singular values of $C$. Take $N$ i.i.d. examples from the distribution and let $\widehat{C}$ denote its sample covariance matrix. Then, if $N \geq K(\sigma_1/\sigma_d^2)R^2 \log(1/\delta)$ the matrix $\widehat{C}$ has full rank with probability at least $1-\delta$. Here $K$ is a universal constant.*

Consider the prefixes produced by Algorithm 1 on input an i.i.d. random sample $\mathcal{S} = (x^1, \ldots, x^N)$ drawn from $\mathcal{D}$. We write $\mathcal{U} = (u^1, \ldots, u^N)$ for the *tuple* of prefixes produced by the algorithm and use $\mathcal{U}'$ to denote the *set* defined by these prefixes. We define $\mathcal{V}$ and $\mathcal{V}'$ similarly. Let $p' = |\mathcal{U}'|$ and $s' = |\mathcal{V}'|$. Our goal is to show that the random sub-block $H' \in \mathbb{R}^{p' \times s'}$ of $H_f$ defined by the output of Algorithm 1 has rank $r$

with high probability w.r.t. the choices of input sample and splitting points. Our strategy will be to show that one always has $H' = P'S'$, where $P' \in \mathbb{R}^{p' \times r}$ and $S' \in \mathbb{R}^{r \times s'}$ are such that with high probability $\mathrm{rank}(P') = \mathrm{rank}(S') = r$. The arguments are identical for $P'$ and $S'$.

Fix a strongly bounded minimal WA $A = \langle \alpha_1^\top, \alpha_\infty, \{A_a\} \rangle$ for $f$, and let $H_f = P_f S_f$ denote the rank factorization induced by $A$. We write $p_u^\top = P_f(u,:)$ for the $u$th row of $P_f$. Note that since $A$ is strongly bounded we have $\|p_u^\top\| = \|\alpha_1^\top A_u\| \leq \|\alpha_1^\top\|$. The desired $P'$ will be the sub-block of $P_f$ correponding to the prefixes in $\mathcal{U}'$. In the following we bound the probability that this matrix is rank deficient.

The first step is to characterize the distribution of the elements of $\mathcal{U}$. Since the prefixes $u^i$ are all i.i.d., we write $\mathcal{D}_p$ to denote the distribution from which these prefixes are drawn, and observe that for any $u \in \Sigma^*$ and any $1 \leq i \leq N$ we have $\mathcal{D}_p(u) = \mathbb{P}[u^i = u] = \mathbb{P}[\exists v : x^i = uv \wedge t = |u|]$, where $x^i$ is drawn from $\mathcal{D}$ and $t$ is uniform in $[0, |x^i|]$. Thus we see that $\mathcal{D}_p(u) = \sum_{v \in \Sigma^*} (1 + |uv|)^{-1} \mathcal{D}(uv)$.

Now we overload our notation and let $\mathcal{D}_p$ also denote the following distribution over $\mathbb{R}^r$ supported on the set of all rows of $P_f$: $\mathcal{D}_p(q^\top) = \sum_{u:p_u^\top = q^\top} \mathcal{D}_p(u)$. It follows from this definition that the covariance matrix of $\mathcal{D}_p$ satisfies $C_p = \mathbb{E}[qq^\top] = \sum_u \mathcal{D}_p(u) p_u p_u^\top$. Observe that this expression can be written in matrix form as $C_p = P_f^\top D_p P_f$, where $D_p$ is a bi-infinite diagonal matrix with entries $D_p(u,u) = \mathcal{D}_p(u)$. We say that the distribution $\mathcal{D}$ is *pref-adversarial* for $A$ if $\mathrm{rank}(C_p) < r$. Note that if $\mathcal{D}(x) > 0$ for all $x \in \Sigma^*$, then $D_p$ has full-rank and consequently $\mathrm{rank}(C_p) = r$. This shows that distributions with full support are never pref-adversarial, and thus we can assume that $C_p$ has full rank.

Next we use the prefixes in $\mathcal{U}$ to build a matrix $P \in \mathbb{R}^{N \times r}$ whose $i$th row corresponds to the $u^i$th row of $P_f$, that is: $P(i,:) = p_{u^i}^\top$. It is immediate to see that $P'$ can be obtained from $P$ by possibly removing some repeated rows and reordering the remaining ones. Thus we have $\mathrm{rank}(P) = \mathrm{rank}(P')$. Furthermore, by construction we have that $\widehat{C}_p = P^\top P$ is the sample covariance matrix of $N$ vectors in $\mathbb{R}^r$ drawn i.i.d. from $\mathcal{D}_p$. Therefore, a straightforward application of Theorem 8 shows that if $N \geq K(\kappa(C_p)/\sigma(C_p))\|\alpha_1^\top\|^2 \log(1/\delta)$, then $\mathrm{rank}(P') = r$ with probability at least $1 - \delta$. Here $K$ is a universal constant, $\kappa(C_p)$ is the condition number of $C_p$, and $\sigma(C_p)$ is the smallest singular value of $C_p$, where these last two terms depend on $A$ and $\mathcal{D}$.

The result follows by symmetry from a union bound. Furthermore, we can take $\eta = \eta(f, \mathcal{D}) = \inf_A \max\{(\kappa(C_p)/\sigma(C_p))\|\alpha_1^\top\|^2, (\kappa(C_s)/\sigma(C_s))\|\alpha_\infty\|^2\}$, where the infimum is taken over all minimal strongly bounded WA for $f$.

## Acknowledgments

This work was partially supported by the EU PASCAL2 NoE (FP7-ICT-216886), and by a Google Research Award. B.B. was supported by an FPU fellowship (AP2008-02064) of the Spanish Ministry of Education. A.Q. and X.C. were supported by the Spanish Government (JCI-2009-04240, RYC-2008-02223) and the E.C. (XLike FP7-288342).

## References

Bailly, R. Quadratic weighted automata: Spectral algorithm and likelihood maximization. *Journal of Machine Learning Research*, 2011.

Balle, B., Quattoni, A., and Carreras, X. A spectral learning algorithm for finite state transducers. *ECML–PKDD*, 2011.

Beimel, A., Bergadano, F., Bshouty, N.H., Kushilevitz, E., and Varricchio, S. Learning functions represented as multiplicity automata. *JACM*, 2000.

Boots, B., Siddiqi, S., and Gordon, G. Closing the learning planning loop with predictive state representations. *I. J. Robotic Research*, 2011.

Dempster, A. P., Laird, N. M., and Rubin, D. B. Maximum likelihood from incomplete data via the EM algorithm. *Journal of the Royal Statistical Society*, 1977.

Hsu, D., Kakade, S. M., and Zhang, T. A spectral algorithm for learning hidden markov models. In *Proc. of COLT*, 2009.

Liu, G., Sun, J., and Yan, S. Closed-form solutions to a category of nuclear norm minimization problems. *NIPS Workshop on Low-Rank Methods for Large-Scale Machine Learning*, 2010.

Luque, F.M., Quattoni, A., Balle, B., and Carreras, X. Spectral learning in non-deterministic dependency parsing. *EACL*, 2012.

Parikh, A.P., Song, L., and Xing, E.P. A spectral algorithm for latent tree graphical models. *ICML*, 2011.

Siddiqi, S.M., Boots, B., and Gordon, G.J. Reduced-rank hidden markov models. *AISTATS*, 2010.

Song, L., Boots, B., Siddiqi, S., Gordon, G., and Smola, A. Hilbert space embeddings of hidden markov models. *ICML*, 2010.

Vershynin, R. *Introduction to the non-asymptotic analysis of random matrices*, volume Compressed Sensing, Theory and Applications, chapter 5. CUP, 2012.